\documentclass[1p,12pt]{elsarticle}

\makeatletter
\def\ps@pprintTitle{
  \let\@oddhead\@empty
  \let\@evenhead\@empty
  \let\@evenfoot\@oddfoot
}
\makeatother

\usepackage[english]{babel}
\usepackage{nicefrac}
\usepackage{amssymb,amsmath,amsthm}
\usepackage[latin1]{inputenc}
\usepackage{eurosym,enumitem}
\usepackage[overload]{empheq}
\usepackage{tikz}
\usepackage{xcolor}
\usepackage{color}
\usepackage{soul}
\usepackage{comment}
\usepackage{eucal}
\usepackage[font=small,skip=9pt]{caption}

\newcommand{\adjoint}{\mathop{\&}\nolimits}

\newcommand{\G}{\text{G}}

\journal{Fuzzy Sets and Systems}

\newtheorem{theorem}{Theorem}
\newtheorem{corollary}[theorem]{Corollary}

\newtheorem{proposition}[theorem]{Proposition}
\newdefinition{definition}[theorem]{Definition }
\newdefinition{remark}[theorem]{Remark}
\newdefinition{example}[theorem]{Example}

\newcommand{\uptriangarrow}{\fontsize{3.5}{1} \kern 1.5 pt \raisebox{-1.75 pt}{\rotatebox{10}{$\nwarrow$}}\normalsize\kern -7 pt\text{{\raisebox{1 pt}{$\vartriangleleft$}}}}
\newcommand{\downtriangarrow}{\fontsize{3.5}{1} \kern 1.5 pt \raisebox{5.75 pt}{\rotatebox{-10}{$\swarrow$}}\normalsize\kern -7 pt\text{{\raisebox{0 pt}{$\vartriangleleft$}}}}
\renewcommand{\qedsymbol}{\fontsize{7}{1} $\blacksquare$ \normalsize}
\newcommand{\Pre}{\text{Pre}}

\begin{document}

\pagenumbering{arabic}

\begin{frontmatter}

\title{Reducing fuzzy relation equations via concept lattices}

\author
{David Lobo, Víctor López-Marchante,  Jes\'{u}s Medina}

\address
{Department of Mathematics, University of  C\'adiz. Spain\\
\texttt{\{david.lobo, victor.lopez, jesus.medina\}@uca.es}
}

\begin{abstract}
This paper has taken into advantage the relationship between Fuzzy Relation Equations (FRE) and Concept Lattices in order to introduce a procedure to reduce a   FRE, without losing information.
Specifically, attribute reduction theory in property-oriented and object-oriented concept lattices has been considered in order to present a mechanism for detecting redundant equations. As a first consequence, the computation of the whole solution set of a solvable FRE is reduced. Moreover, we will also introduce  a novel method for computing approximate solutions of unsolvable FRE  related to a (real) dataset with uncertainty/imprecision data.
 \end{abstract}


\end{frontmatter}

\section{Introduction}\label{sec:introduction}

Fuzzy relation equations (FRE) arise as an approximate reasoning tool for discerning the possible relations masked in a database, which is instrumental in real world problems, as  Sanchez claimed in his seminal papers \cite{sanchez76, sanchez79}: \emph{``Human judgments are often based on comparisons between couples of faced data''}.
Lying their foundations in fuzzy set theory~\cite{Z1}, FRE features as a powerful technique for handling databases containing uncertain and incomplete information.   
   
The applications of FRE cover an ample range of fields as decision making~\cite{Cornejo2017,Fan98,Li2013DM}, bipolarity~\cite{CLM:MMA2019,CLM:JCAM2018,FREStandard:FSS2020,Freson2013}, image processing~\cite{ijuks17:morph,Hirota2002,Loia2005}, fuzzy control~\cite{dinola91,Nola:1989,Feng2013}, optimization~\cite{Chang2013,Ghodousian2008,Mazraeh2018} and abductive reasoning~\cite{dubois1995,CLM:eusflat2019}. Their large usage evince the necessity of solving FRE. Since their introduction, different authors started to study the resolution of FRE with generalized connectives~\cite{belohlavek:2002,baets99_eq,Nola:1989,PedryczGC:83,Peeva2013,Turunen1987}. For example, we can find works as \cite{baets99_eq}, in which the resolution of FRE is based on the so-called ``root systems'', and~\cite{Nola:1989}, where a complete study of FRE is done. On the other hand, there exist diverse developments which relate the resolution of FRE with other frameworks. For instance, the authors in~\cite{Markovskii} associate the study of FRE with the resolution of covering problems, and two researches on approximating solutions of FRE can be found in~\cite{CornejoFRE2017,perfilieva04}. In this paper, we are interested in the existing correspondence between FRE and Formal Concept Analysis (FCA), which was shown by D\'iaz-Moreno and Medina in~\cite{dm:mare,dm:ins2014}.

Formal Concept Analysis {(FCA)}~\cite{GanterW} is a mathematical tool allowing the extraction of information from relational databases (object-attributes) by means of antitone Galois connections. On the other hand, rough set theory is another formal tool for extracting information using isotone Galois connections~\cite{DuntschG03,DuntschG02}. Multi-adjoint property-oriented and object-oriented concept lattices~\cite{mor-fss-cmpi} {merge both} these two theories, following the original idea of D{\"u}ntsch and Gediga~\cite{DuntschG03,DuntschG02}, and Chen and Yao~\cite{chenyao08, yao06}. A common technique used in the FCA framework is the attribute classification and reduction~\cite{Cornejo2017,ar:ins:2015,fuzzieee17:redobject}, which enables to decide whether an attribute contains essential information or not.

The authors in~\cite{dm:mare} proved that the solutions of a FRE and the concepts of an associated multi-adjoint property-oriented concept lattice are related. In other words, there exists a link between the solution set of a FRE and its associated concept lattice. The underlying idea of this paper is applying attribute reduction procedures in order to reduce multi-adjoint fuzzy relation equations, in the sense of eliminating redundant equations, as the Gaussian elimination method does in the usual  systems of linear equations, but in the complex framework of the (multi-adjoint) FRE.

As it is well-known in the literature of FRE, the computation of the solutions of a FRE entails a high cost, as it is a NP-hard problem~\cite{chenwang02}. {As a consequence, the  reduction method proposed in this paper on} multi-adjoint FRE becomes instrumental in what concerns large databases. Furthermore, as it will turn out, the reduction method presented in this paper is only based on the coefficient matrix of the given FRE. Hence, the reduction of FRE shown here plays a key role when having different instances of the same system. For example, if a system corresponds to the coefficient matrix $A$ and there are several different observations $b_1,b_2,\dots,b_n$, we obtain a set of $n$ FRE of the form $A\odot x=b_i$, with $i\in\{1,\dots,n\}$. Now, performing a reduction mechanism on the coefficient matrix $A$, the resolution of all FRE can be simultaneously simplified.

A problem that often arises in real-world situations is the unsolvability of FRE, due to contradictory information. This may be caused by many different reasons, like employing inaccurate measure methods, or simply by the imprecise data usually present in the considered dataset. In this paper, we show how the relation between FRE and concept lattices can be used for slightly modifying the independent term in an unsolvable FRE so that it becomes solvable. Namely, depending on the underlying concept lattice, we may obtain different ways of approximating an unsolvable FRE by a solvable FRE. Despite the fact that in~\cite{CornejoFRE2017} a useful approach to the approximation of unsolvable FRE was developed, the one that will be introduced in this paper grants that only certain values in the independent term will be modified. Indeed, both mechanisms can be combined to capture the main features of each one.

The algebraic structures  considered in this paper are multi-adjoint lattices due to their flexibility, although other structures could be considered as Heyting algebras or general residuated lattices, as we will see at the end of the paper. In the multi-adjoint framework, properties of several different conjunctions can be used. In~\cite{CornejoFRE2017}, the multi-adjoint paradigm was extended to FRE, leading to what was called multi-adjoint relation equations (MARE). These new FRE were used for dealing with multi-adjoint logic programming (MALP) in the computation of weights and abductive reasoning. The applications of MALP have recently been studied in~\cite{SCI23JATZ}, where the effects of choosing different operators in a MALP have been discussed.
 
The paper is organized as follows. In Section~\ref{sec:preliminaries}, some preliminary definitions and results related to FCA and FRE will be recalled. In Section~\ref{sec:reduction}, the notion of reduced FRE is introduced and we show that the consistency of a set of attributes is a sufficient condition for reducing a FRE without loosing essential information, that is, for obtaining a reduced FRE whose solution set coincides with the solution set of the complete FRE. In Section~\ref{sec:approximation}, we present the concept of feasible reduct and we show that every feasible reduct provides a solvable approximation of an unsolvable FRE. Some examples are included to illustrate how the approximation mechanism can be used to identify incoherences in a database. Finally, in Section~\ref{sec:dual} an analogous study is presented with respect to dual FRE basing on object classification and reduction. The paper concludes with some conclusions and prospects for future work.

\section{Preliminaries}\label{sec:preliminaries}

In this section, some preliminary notions and results related to FCA~\cite{GanterW, mor-fss-cmpi} and FRE~\cite{Nola:1989,dm:mare,dm:ins2014} will be recalled.

Adjoint triples~\cite{Cornejo2021} are the basic operators that will be used throughout this paper, being a generalisation of left-continuous t-norms and its residuated implications. 
\begin{definition}[\cite{Cornejo2021}] \label{def:adj_triple}
	Let $(P_1,\preceq_1)$, $(P_2,\preceq_2)$, $(P_3,\preceq_3)$ be posets and $\&\colon P_1\times P_2 \rightarrow P_3$, $\swarrow\colon P_3 \times P_2 \rightarrow P_1$, $\nwarrow\colon P_3\times P_1 \rightarrow P_2$ mappings, then $(\&,\swarrow, \nwarrow)$ is called an \emph{adjoint triple} with respect to $P_1$, $P_2$, $P_3$ if
		$$x\preceq_1 z\swarrow y \hspace{10pt} \mbox{ iff } \hspace{10pt} x\& y \preceq_3 z  \hspace{10pt}\mbox{ iff } \hspace{10pt} y\preceq_2 z\nwarrow x$$
	for each $x\in P_1$, $y\in P_2$, $z\in P_3$.
\end{definition}
Following the multi-adjoint philosophy, the basic algebraic structure here will be a triplet of posets endowed with a set of adjoint triples. For computational reasons, we demand two of the posets to be lattices.
\begin{definition}[\cite{ins-medina}] \label{def:MA_frame} Let $(L_1,\preceq_1)$, $(L_2,\preceq_2)$ be two lattices, $(P,\leq)$ a poset and $\{(\&_i,\swarrow^i, \nwarrow_i)\mid i\in \{1,\dots,n\}\}$ a set of adjoint triples with respect to $P, L_2, L_1$. The tuple $$(L_1, L_2, P, \preceq_1,\preceq_2,\preceq, \&_1, \swarrow^1,\nwarrow_1,\dots, \&_n, \swarrow^n,\nwarrow_n)$$ is called \emph{multi-adjoint property-oriented frame}.
\end{definition}
To improve readability, if $L_1=L_2=P$, the property-oriented multi-adjoint frame will be denoted as $$(P, \preceq, \&_1, \swarrow^1,\nwarrow_1,\dots, \&_n, \swarrow^n,\nwarrow_n)$$
The reality under study is represented by the formal notion of context. Basically, a context consists of a set of objects, a set of attributes, a relation between them and a mapping that assigns (the index of) an adjoint triple to each pair attribute-object.

\begin{definition}[\cite{ins-medina}]\label{def:context} Let $(L_1, L_2, P, \preceq_1,\preceq_2,\preceq, \&_1, \swarrow^1,\nwarrow_1,\dots, \&_n, \swarrow^n,\nwarrow_n)$ be a property-oriented multi-adjoint frame. A \emph{context} is a tuple $(A,B,R,\sigma)$ where $A$ and $B$ are non-empty sets, $R\colon A\times B\rightarrow P$ is a fuzzy relation and $\sigma\colon A\times B\rightarrow\{1,...,n\}$ is a mapping.
\end{definition}

In the definition of multi-adjoint context, the set $A$ is usually interpreted as the set of attributes and the set $B$, as the set of objects. Hence, the mapping $\sigma$ assigns (the index of) an adjoint triple to each pair of attribute and object. From now on, we will fix a multi-adjoint property-oriented frame $(L_1, L_2, P, \preceq_1,\preceq_2,\preceq, \&_1, \swarrow^1,\nwarrow_1,\dots, \&_n, \swarrow^n,\nwarrow_n)$ and a context $(A,B,R,\sigma)$. Besides, to improve readability, we will write $\&_{a,b}$ instead of $\&_{\sigma(a,b)}$.

In this environment, considering the fuzzy subsets of attributes $L_1^{A}=\{f\mid f\colon A\rightarrow L_1\}$ and the fuzzy subsets of objects $L_2^{B}=\{g\mid g\colon B\rightarrow L_2\}$, the mappings $^{\uparrow_\pi}\colon L_2^B \rightarrow L_1^A$ and $^{\downarrow^N}\colon L_1^A \rightarrow L_2^B$ are defined as follows
\begin{align}
	g^{\uparrow_\pi}(a)&=\bigvee
	\raisebox{-5pt}{\kern -2 pt $\scriptstyle1$}
	\left\lbrace R(a,b)\hspace{3 pt}\&_{a,b}\hspace{3 pt}g(b)\mid b\in B\right\rbrace \label{exp:possibility}\\
	f^{\downarrow^N}(b)&=\bigwedge
	\raisebox{-5pt}{\kern -1 pt $\scriptstyle 2$}
	\left\lbrace f(a)\nwarrow_{a,b}R(a,b)\mid a\in A\right\rbrace \label{exp:necesity}
\end{align}
for each $f\in L_1^A$ and $g\in L_2^B$, where $\bigvee \raisebox{-5pt}{\kern -2 pt $\scriptstyle 1$}
$ and $\bigwedge
\raisebox{-5pt}{\kern -1 pt $\scriptstyle 2$}
$ represent the supremum and infimum of the lattices $(L_1,\preceq_1)$ and $(L_2,\preceq_2)$, respectively. Among other properties, the pair $(^{\uparrow_\pi},^{\downarrow^N}\kern -3.5 pt)$ verifies to be an isotone Galois connection~\cite{ins-medina}. This leads to the definition of \emph{multi-adjoint property-oriented concept lattice}. Namely, consider the order relation $\preceq_{\pi N}$ defined as $(g_1,f_1)\preceq_{\pi N}(g_2,f_2)$ if and only if $f_1\preceq_1 f_2$, or equivalently, if and only if $g_1\preceq_2 g_2$. The multi-adjoint property-oriented concept lattice associated with the multi-adjoint property-oriented frame is given by
\begin{equation}\label{exp:clattice}
	\CMcal{M}_{\pi N}(A,B,R,\sigma)=\left\lbrace (g,f)\in L_2^B\times L_1^A\mid g=f^{\downarrow^N},f=g^{\uparrow_\pi}  \right\rbrace
\end{equation}  
As shown in~\cite{ins-medina}, the set $\CMcal{M}_{\pi N}$ together with the order $\preceq_{\pi N}$ forms a complete lattice. Given a concept $(g,f)$, $g$ is called \emph{extent} of the concept and $f$ is called \emph{intent} of the concept. The set of all extents will be denoted as $\CMcal{E}(\CMcal{M}_{\pi N})$ and the set of all intents as $\CMcal{I}(\CMcal{M}_{\pi N})$. Both sets have the algebraic structure of complete lattice with the order that $\preceq_{\pi N}$ induces on each component.

The previous notions admit a dual version, giving rise to the concepts of \textit{multi-adjoint object-oriented frame} and \textit{multi-adjoint object-oriented concept lattice}. For more details, we refer the reader to~\cite{ins-medina}.

The following definitions aim to reduce large contexts in order to improve their handleability. The underlying idea is reducing the set of attributes, {whilst loosing as little information as} possible. First and foremost, it is necessary to establish when two concept lattices are isomorphic with respect to their extents, because it will be enable us to know whether the information under the dataset changes when removing an attribute.

This notion will be strongly related to the notion of isomorphism between ordered sets. Two posets $(P_1, \preceq_1), (P_2, \preceq_2)$ are isomorphic if there exists a bijective mapping $\phi\colon P_1\rightarrow P_2$ such that, for all $x\in P_1$ and $y\in P_2$, $x\preceq_1 y$ if and only if $\phi(x)\preceq_2 \phi(y)$. Taking into account that the set of extents of a concept lattice is a complete lattice, the notion of $\CMcal{E}$-isomorphism follows easily, where $\CMcal{E}$ denotes that the sets of extents are isomorphic.

\begin{definition}[\cite{DaveyPriestley}]\label{def:E-isomorphism} Let $\CMcal{M}_1$, $\CMcal{M}_2$ be concept lattices. We say that $M_1$ and $M_2$ are \emph{isomorphic with respect to their extents, or $\CMcal{E}$-isomorphic}, denoted as $\CMcal{M}_1 \cong_\CMcal{E}\CMcal{M}_2$, if their sets of extents $\CMcal{E}(\CMcal{M}_1)$, $\CMcal{E}(\CMcal{M}_2)$ are isomorphic.
\end{definition}

When reducing a context, it is pretended that the associated concept lattice remains the same, because the most relevant information is contained in it. Hence, we will look for sets of attributes generating an isomorphic concept lattice with respect to the complete/original one, what means that the set of attributes contains the essential information.

\begin{definition}[\cite{ar:ins:2015}]\label{def:E-consistent}
	A set of attributes $Y\subseteq A$ is an \emph{$\CMcal{E}$-consistent set of $(A,B,R,\sigma)$} if the following isomorphism holds:
	$$\CMcal{M}_{\pi N}(Y,B,R_Y,\sigma_{Y\times B})\cong_\CMcal{E}\CMcal{M}_{\pi N}(A,B,R,\sigma)$$	
Where $R_Y$ and $\sigma_{Y\times B}$ represent the restrictions of the fuzzy relation $R$ and the mapping $\sigma$ to the sets $Y$ and $Y\times B$, respectively.
\end{definition}

Notice that, the conditions of Definition~\ref{def:E-consistent} are equivalent to say that, for all $(g,f)\in\CMcal{M}_{\pi N}(A,B,R,\sigma)$, there exists $(g',f')\in \CMcal{M}_{\pi N}(Y,B,R_Y,\sigma_{Y\times B})$ such that $g=g'$.

Manipulating contexts with many attributes and objects requires a large number of operations. When dealing with a reduced context, for the sake of shortening computation times, it is natural that we want an $\CMcal{E}$-consistent set of attributes to have as few elements as possible, what leads us to the next definition: the minimal sets of attributes containing the essential information.

\begin{definition}[\cite{ar:ins:2015}] \label{def:E-reduct}
	A set of attributes $Y\subseteq A$ is an $\CMcal{E}$-reduct of $(A,B,R,\sigma)$ if it is a consistent set of $(A,B,R,\sigma)$ and, for all $a\in Y$, $$\CMcal{M}_{\pi N}(Y\backslash \{a\},B,R_{Y\backslash \{a\}},\sigma_{Y\backslash \{a\}\times B})\ncong_\CMcal{E}\CMcal{M}_{\pi N}(A,B,R,\sigma)$$
\end{definition}

Reasoning analogously for a multi-adjoint object-oriented frame, Definitions~\ref{def:E-consistent} and~\ref{def:E-reduct} lead to object-reduction. In that case, the notions of isomorphism and reduct are based on the intents of the concept lattice, rather than on the extents, from which the terms $\CMcal{I}$-reduct and $\CMcal{I}$-isomorphism are defined.

From now on, in order to improve readability, $\CMcal{E}$-consistent sets will be simply called consistent sets and $\CMcal{E}$-reducts will be called reducts when there is no room for confusion.

 In what follows, we recall some notions and results concerning FRE. For more details, we refer the reader to~\cite{dm:mare}.
 
 A multi-adjoint FRE is a problem of the form $R\odot S=T$, where $R,S$ and $T$ are fuzzy relations, $\odot$ is a composition operator and $R$ or $S$ is an unknown relation. Naturally, the definition of $\odot$ is relevant in the solvability of the problem. In this approach, two different compositions will be defined, both of them based on the multi-adjoint paradigm. A third one can be defined by using the implication $\swarrow$, but we will omit its formal definition here.

\begin{definition}[\cite{dm:mare}]\label{def:compositions}
	Let $U,V,W$ be sets, $(P_1,\preceq_1),(P_2,\preceq_2),(P_3,\preceq_3)$\linebreak posets, $\{(\&_i,\swarrow^i,\nwarrow_i)\mid i\in \{1,\dots, n\}\}$ a set of adjoint triples with respect $P_1,P_2,P_3$, $\sigma\colon V\rightarrow \{1,\dots, n\}$ a mapping, and $R\in P_1^{U\times V}$, $S\in P_2^{V\times W}$, $T\in P_3^{U\times W}$ three fuzzy relations.
	\begin{enumerate}
		\item If $P_3$ is a complete lattice, the operator $\odot_\sigma\colon P_1^{U\times V}\times P_2^{V\times W} \rightarrow P_3^{U\times W}$ defined as
		\begin{equation}\label{exp:supcomp}R\odot_\sigma S (u,w)=\bigvee
		\raisebox{-5pt}{\kern -2 pt$\scriptstyle 3$}
		\{R(u,v)\&_{\sigma(v)}S(v,w)\mid v\in V\}\end{equation}
		is called \emph{sup-$\&_\sigma$-composition}.
		\item If $P_2$ is a complete lattice, the operator $\uptriangarrow_\sigma\colon P_3^{U\times W} \times P_1^{U\times V}\rightarrow P_2^{V\times W}$ defined as
		\begin{equation}\label{exp:infnwcomp}T\uptriangarrow_\sigma R (v,w)=\bigwedge
			\raisebox{-5pt}{\kern -1 pt$\scriptstyle 2$}
		\{T(u,w)\nwarrow_{\sigma(v)}R(u,v)\mid u\in U\}\end{equation}
		is called \emph{inf-$\nwarrow_\sigma$-composition}.
	\end{enumerate} 
\end{definition}

Notice that, the completeness of lattices is enough for ensuring that infima and suprema in expressions \eqref{exp:supcomp} and \eqref{exp:infnwcomp} are well defined. Furthermore, it is sufficient to define them in an upper semilattice or a lower semilattice, respectively.

Compositions $\odot_\sigma$ and $\uptriangarrow_\sigma$ lead to different FRE. In what follows, we will fix the framework considered in Definition~\ref{def:compositions} and we will assume that $P_3$ is a complete lattice.

\begin{definition}[\cite{dm:mare}]\label{def:equations}
	A multi-adjoint FRE with sup-$\&_\sigma$-composition is an equality of the form
	\begin{equation}\label{exp:eq1}
		R\odot_\sigma X=T
	\end{equation}	
	or of the form
	\begin{equation}\label{exp:eq2}
		X\odot_\sigma S=T
	\end{equation}
	where, in both cases, $X$ is an unknown fuzzy relation.
	 
	 Notice that the equation $R\odot_\sigma X=T$ can clearly be written as different  systems of equations. Specifically, for each $w\in W$, we have the following system
	\begin{equation}\label{sys:FRE_general}
	\begin{tabular}{ccccccc}
	$R(u_1,v_1)\,\&_{\sigma(v_1)}\,x_1$	& $\vee$ & $\cdots$ & $\vee$ & $R(u_1,v_m)\,\&_{\sigma(v_m)}\, x_m$ & $=$ & $t_1$ \\
	$\vdots$	&  & $\vdots$ &  & $\vdots$  &  & $\vdots$  \\
	$R(u_n,v_1)\,\&_{\sigma(v_1)}\, x_1$ & $\vee$	& $\cdots$  & $\vee$ & $R(u_n,v_m)\,\&_{\sigma(v_m)}\, x_m$ & = & $t_n$ \\
	\end{tabular}
	\end{equation}
where   $X(v_j,w)=x_{j}$  $T(u_i,w)=t_{i}$, for all $i\in\{1,\ldots,n\}$, $j\in\{1,\ldots,m\}$.

We say that a multi-adjoint FRE is \emph{solvable} if there exists at least one solution, that is, a relation $X$ satisfying \eqref{exp:eq1} or \eqref{exp:eq2}. Otherwise, we say it is \emph{unsolvable}.
\end{definition}

The remaining notions and results of the preliminaries section are presented with respect to Equation~\eqref{exp:eq1}. Nevertheless, a dual version of these is obtained concerning Equation~\eqref{exp:eq2}, as shown in~\cite{dm:mare, dm:ins2014}. The next definition associates a context with a given FRE and, in consequence, with a concept lattice.

\begin{definition}[\cite{dm:mare}]\label{def:asoc_context}
	Consider a MARE $R\odot_\sigma X=T$ where $R\colon U\times V \to P$, $X\colon V\times W\to L_1$, $T\colon U\times W\to L_1$ are fuzzy relations and $\sigma\colon U\times V\to \{1,\dots,n\}$ is the mapping used to define the sup-composition operator $\odot_\sigma$. The \emph{multi-adjoint context associated with $R\odot_\sigma X=T$} is $(U,V,R,\sigma)$.
\end{definition}

A necessary and sufficient condition for a relation $R$ satisfying Equation~\eqref{exp:eq1} is provided below in terms of the Galois connection $(^{\uparrow_\pi},^{\downarrow^N}\kern -3.5 pt)$ defined on the associated context $(U,V,R,\sigma)$.

\begin{proposition}[\cite{dm:mare}]\label{prop:charact_sol}
	Let $(U,V,R,\sigma)$ be the multi-adjoint context associated with the multi-adjoint FRE $R\odot_\sigma X=T$. A fuzzy relation $X\in L_2^{V\times W}$ is a solution of the FRE if and only if we obtain that
	$$X_w^{\uparrow_\pi}=T_w$$
for all $w\in W$, where $X_w$ and $T_w$ are the columns of $X$ and $T$, respectively, that is, $X_w(v)=X(v,w)$ and $T_w(u)=T(u,w)$, for all $u\in U$, $v\in V$ and $w\in W$.
\end{proposition}

In particular, Proposition~\ref{prop:charact_sol} simplifies the process of checking whether a fuzzy relation is a solution of a multi-adjoint FRE or not.	
The following result characterizes the solvability of a multi-adjoint FRE by means of the concept lattice associated with the FRE. Additionally, in case of being solvable, it provides the maximum solution of the FRE.

\begin{proposition}[\cite{dm:mare}]\label{prop:charact_solvable}
		Let $(U,V,R,\sigma)$ be the multi-adjoint context associated with a  {multi-adjoint} FRE $R\odot_\sigma X=T$ and $(\CMcal{M}_{\pi N},\preceq_{\pi N})$ the concept lattice associated with that context. Then $R\odot_\sigma X=T$ is solvable if and only if $T_w\in \CMcal{I}(\CMcal{M}_{\pi N})$ for all $w\in W$. {In that case}, $T\uptriangarrow_\sigma R\in L_2^{V\times W}$ is the maximum solution.
\end{proposition}
Notice that, the condition  $T_w\in \CMcal{I}(\CMcal{M}_{\pi N})$ is equivalent to  $T_w^{\downarrow^N\uparrow_\pi}=T_w$. The following corollary provides, if exists, a direct way of computing the maximum solution of a FRE.
\begin{corollary}[\cite{dm:mare}]\label{cor:max_sol} 
	Let $(U,V,R,\sigma)$ be the multi-adjoint context associated with a solvable multi-adjoint FRE $R\odot_\sigma X=T$ and $(\CMcal{M}_{\pi N},\preceq_{\pi N})$ the concept lattice associated with that context. The maximum solution verifies that $(T\uptriangarrow_\sigma R)_w=T_w^{\downarrow^N}$ for each $w\in W$.
\end{corollary}
	
Last but not least, the following result was proved in~\cite{dm:ins2014} to characterize the whole solution set of a multi-adjoint FRE.

\begin{proposition}[\cite{dm:ins2014}]\label{prop:charact_predec}
		Let $(U,V,R,\sigma)$ be the multi-adjoint context associated with a multi-adjoint FRE $R\odot_\sigma X=T$ and $(\CMcal{M}_{\pi N},\preceq_{\pi N})$ the concept lattice associated with that context. Given $X_w\in L_2^{V}$ with $w\in W$, then $X_w^{\uparrow_\pi}=T_w$ if and only if $X_w\preceq_2 T_w^{\downarrow^N}$ and there is no $(g,f)\in \CMcal{M}_{\pi N}$ such that $X_w\preceq_2 g\prec_2 T_w^{\downarrow^N}$.
\end{proposition}
Given a lattice $(L,\preceq)$ and $x\in L$, we will denote the lower bounds of $x$ as $(x]$. Additionally, we define the set of predecessors of $x$ as
		$$\Pre_L(x)=\{x'\in L\mid x'\prec x\mbox{ and there is no } x''\in L \mbox{ such that } x'\prec x''\prec x\}$$
		
		The solution set of a FRE can be characterized as follows.

\begin{corollary}[\cite{dm:ins2014}]\label{cor:charact_sol_set}
	Let $(U,V,R,\sigma)$  be the multi-adjoint context associated with a solvable multi-adjoint FRE $R\odot_\sigma X=T$. If $U,V$ are finite sets, the solution set of the FRE is the following
	\[
		\left\lbrace X\in L_2^{V\times W}\mid X_w\in  \left( T_w^{\downarrow^N}\right] 
	\Big\backslash
	\bigcup \left\lbrace  \left( g\right]  \mid g\in \Pre_{\CMcal{E}(\CMcal{M}_{\pi N})}\left( T_w^{\downarrow^N}\right) \right\rbrace \mbox{ for each }w\in W\right \rbrace
	\]
\end{corollary}

\section{Multi-adjoint FRE reduction}\label{sec:reduction}
The procedures given in the {previous} section enable to study the solvability of a multi-adjoint FRE and, in case it is solvable, to compute the whole solution set. 
In this section, we will study whether it is possible to simplify the number of equations of the systems associated with a FRE (System~\eqref{sys:FRE_general}), that is, whether redundant equations exist in  a multi-adjoint FRE $R\odot_\sigma X=T$. As a consequence of this study, the computation of the whole solution set of a (multi-adjoint) FRE can be reduced. Namely, the number of operations needed for such computation can be too large in certain circumstances, what can be a problem when the database that is under study has a high number of elements. For example, lets fix a multi-adjoint frame
\begin{equation}\label{exp:MA_frame_fixed} 
([0,1]_n, \leq, \&_1, \swarrow^1,\nwarrow_1,\dots, \&_n, \swarrow^n,\nwarrow_n)
\end{equation}
where $[0,1]_n$ denotes the regular partition of the unit interval in $n$ pieces.
Consider the sets 
$$
U=\{u_i\mid i\in \{1,\dots,m\}\}, V=\{v_i\mid i\in \{1,\dots,m\}\}, W=\{w\}$$ and  {a solvable} {multi-adjoint FRE} 
$$
R\odot_\sigma X=T
$$
 where $R\in [0,1]_n^{U\times V}$, $T\in [0,1]_n^{U\times W}$ and $X\in [0,1]_n^{V\times W}$, being $X$ unknown.

Notice that, by Corollary~\ref{cor:max_sol}, it is easy to obtain the maximum solution of the equation. Nevertheless, the procedure for describing the whole solution set would require, in the worst case, computing the whole concept lattice associated with it. Such computation has a very high computational cost due to the large number of operations involved in it. For example, the number of candidates for being the intent of a concept is $(n+1)^m$ (each one of the fuzzy subsets of $[0,1]_n^U$). Although some procedures can optimize the computation \cite{BELOHLAVEK2019132}, the number of operations scales quickly, so with a short number of variables, the time of computation becomes excessive.

The situation changes if we remove some elements from $U$, since the number of possibilities is clearly smaller. Observe that, in order to ensure that the solution set of the reduced equation is the same as the original one, this reduction should only remove  redundant elements.

Therefore, in this line, we will study the impact of the attribute reduction theory, developed in the (multi-adjoint) FCA framework, to the reduction of redundant equations of a given FRE. As a result, we will show that the consideration of a consistent set of attributes will allow us to reduce the equation preserving the original information. By contrary, if the considered set  is not consistent, the reduced system may miss some information, which would lead to dissimilarities between the solution set of the reduced system and the solution set of the complete one. For example, the solutions of the reduced system could not necessarily be solutions of the complete equation.

In what follows, three finite sets $U$,$V$,$W$ and a multi-adjoint property-oriented frame $(L_1, L_2, P, \preceq_1,\preceq_2,\preceq, \&_1, \swarrow^1,\nwarrow_1,\dots, \&_n, \swarrow^n,\nwarrow_n)$, will be fixed. Consider a FRE {$R\odot_{\sigma}X=T$}, where $R\in P^{U\times V}$, $T\in L_1^{U\times W}$ and $X\in L_2^{V\times W}$, being $X$ unknown, and let $(U,V,R,\sigma)$ be its associated multi-adjoint context. The following notion formalizes the reduction of a FRE by removing elements in the set~$U$.

\begin{definition}\label{def:reduc_FRE}
	Let $Y\subseteq U$ and consider the relations $R_Y=R_{\mid Y\times V}$, $T_Y=T_{\mid Y\times W}$. The multi-adjoint FRE ${R_Y}\odot_{\sigma}X={T_Y}$ is called \emph{$Y$-reduced FRE of $R\odot_{\sigma}X=T$}.
\end{definition}

Notice that, the elimination of elements in $U$ implies the reduction of rows in $R$ and $T$, which entails the elimination of equations of the system associated with $R\odot_{\sigma}X=T$ (System~\eqref{sys:FRE_general}).

Given a subset of attributes $Y\subseteq U$, we will use $(^{\uparrow_\pi^Y},^{\downarrow^N_Y}\kern -2.5 pt)$ to denote the Galois connection associated with the context $(Y,V,R_Y,\sigma_{\mid Y\times V})$.

\begin{remark}\label{rem:possibility}
	Let $g\in L_2^B$ and $Y\subseteq U$. By definition of the operator $^{\uparrow_\pi}$, the mapping $g^{\uparrow_\pi}$ is defined, for each $u\in U$, as 
	$$g^{\uparrow_\pi}(u)=\bigvee
	\raisebox{-5pt}{\kern -2 pt $\scriptstyle 1$}
	\{R(u,v)\hspace{3 pt}\&_{u,v}\hspace{3 pt}g(v)\mid v\in V\}$$
	Analogously, the mapping $g^{\uparrow_\pi^Y}\in L_1^Y$, is defined for each $y\in Y$ as
	$$g^{\uparrow_\pi^Y}(y)=\bigvee
	\raisebox{-5pt}{\kern -2 pt $\scriptstyle 1$}
	\{R(y,v)\hspace{3 pt}\&_{y,v}\hspace{3 pt}g(v)\mid v\in V\}$$
	so for each $y\in Y$, it is satisfied that $g{^{\uparrow_\pi^Y}}(y)=g{^{\uparrow_\pi}}(y)$.
\end{remark}

From here on, for each $Y\subseteq U$, the concept lattice $\CMcal{M}_{\pi N}(Y,V,R_Y,\sigma_{\mid Y\times V})$ will be denoted as $\CMcal{M}_{\pi N}(Y)$ to improve readability.

The following result shows that, if a multi-adjoint FRE is solvable and $Y$ is consistent, then a relation is a solution of the $Y$-reduced version of the FRE if and only if it is a solution of the complete FRE. In particular, this implies that the solution set of any reduced FRE with respect to a consistent set coincides with the solution set of the complete FRE.

\begin{theorem}\label{th:red-cons}
Let $R\odot_{\sigma}X=T$ be a solvable FRE and $Y$ a consistent set of $(U,V,R,\sigma)$. We have that the $Y$-reduced FRE of $R\odot_{\sigma}X=T$ is solvable. Moreover, $\overline{X}\in L_2^{V\times W}$ is a solution of the $Y$-reduced FRE if and only if it is a solution of the complete FRE.
\end{theorem}
\begin{proof}	
Let $(\CMcal{M}_{\pi N}(U),\preceq_{\pi N})$ and $(\CMcal{M}_{\pi N}(Y),\preceq_{\pi N})$ be the associated concept lattices of the complete FRE and the $Y$-reduced FRE, respectively. Applying Proposition~\ref{prop:charact_solvable}, we will see that $R_Y\odot_{\sigma}X=T_Y$ is solvable by proving that $(T_Y)_w\in \CMcal{I}(\CMcal{M}_{\pi N}(Y))$ for all $w\in W$.
	
	Consider $w\in W$ fixed. Since $R\odot_{\sigma}X=T$ is solvable, by Proposition~\ref{prop:charact_solvable}, $T_w\in \CMcal{I}(\CMcal{M}_{\pi N}(U))$. Hence,  $T_w^{\downarrow^N}\in \CMcal{E}(\CMcal{M}_{\pi N}(U))$. Notice that, as $Y$ is a consistent set, {$\CMcal{M}_{\pi N}(U)\cong_{\CMcal{E}} \CMcal{M}_{\pi N}(Y)$}, from which, $T_w^{\downarrow^N}\in \CMcal{E}(\CMcal{M}_{\pi N}(Y))$. Therefore, there exists a concept in $\CMcal{M}_{\pi N}(Y)$ of the form $(T_w^{\downarrow^N},f_w)$. In what follows, we will see that $f_w=(T_Y)_w$ and, as a consequence, $(T_Y)_w\in\CMcal{I}(\CMcal{M}_{\pi N}(Y))$, which completes the first part of the proof.
	 
	Because $(T_w^{\downarrow^N},f_w)\in\CMcal{M}_{\pi N}(Y)$, then $f_w=(T_w^{\downarrow^N}){^{\uparrow_\pi^Y}}$.
	In particular, applying Remark~\ref{rem:possibility}, for each $y\in Y$ it holds that $$f_w(y)={(T_w^{\downarrow^N})^{\uparrow_\pi^Y}(y)=T_w^{\downarrow^N{\uparrow_\pi}}(y)}$$
	Now, since $T_w\in \CMcal{I}(\CMcal{M}_{\pi N}(U))$, then $T_w^{\downarrow^N{\uparrow_\pi}}=T_w$. Furthermore, by definition of restriction, $T_w(y)=(T_Y)_w(y)$ for all $y\in Y$. Consequently, the following chain of equalities holds, for all $y\in Y$,
	$$f_w(y)=T_w^{\downarrow^N{\uparrow_\pi}}(y)=T_w(y)=(T_Y)_w(y)$$
	
	Hence, as $(T_Y)_w=f_w\in \CMcal{I}(\CMcal{M}_{\pi N}(Y))$ for all $w\in W$, we conclude from Proposition~\ref{prop:charact_solvable} that ${R_Y}\odot_{\sigma}X=T_Y$ is solvable.
	
	We will now see that a fuzzy relation is a solution of the $Y$-reduced FRE if and only if it is a solution of the complete FRE.
	
	Let $\overline{X}$ be a solution of the {complete} FRE. Equivalently, by Proposition~\ref{prop:charact_sol}, $\overline{X}_w^{\uparrow_\pi}=T_w$ for all $w\in W$. On the one hand, applying Remark~\ref{rem:possibility}, $\overline{X}_w^{\uparrow_\pi}(y)=\overline{X}_w^{\uparrow_\pi^Y}(y)$ for all $y\in Y$. On the other hand, due to the definition of restriction, we have $T_w(y)= (T_Y)_w(y)$ for all $y\in Y$. Therefore, the following chain of equalities holds for all $y\in Y$ and $w\in W$
	$$\overline{X}_w^{\uparrow_\pi^Y}(y)=\overline{X}_w^{\uparrow_\pi}(y)=T_w(y)=(T_Y)_w(y)$$
	As a result, by Proposition~\ref{prop:charact_sol}, $\overline{X}$ is a solution of the $Y$-reduced FRE.
	
	Now, let $\overline{X}$ be a solution of the $Y$-reduced FRE and let us see that it is also a solution of the complete FRE. Equivalently, by Proposition~\ref{prop:charact_sol}, we will prove that $\overline{X}_w^{\uparrow_\pi}=T_w$ for all $w\in W$. For this purpose, we will check that $\overline{X_w}^{{\uparrow_\pi\downarrow^N}}=T_w^{\downarrow^N}$ for all $w\in W$. Then, applying $^{\uparrow_\pi}$ on both sides of the equality, we will obtain the pursued expression.	
	
	Consider $w\in W$ fixed. Applying Proposition~\ref{prop:charact_sol}, since $\overline{X}$ is solution of the $Y$-reduced FRE, $\overline{X}_w^{{\uparrow_\pi^Y}}=(T_Y)_w$. Now, applying $^{{\downarrow^N_Y}}$ on both sides of the equality and using that $^{{\uparrow_\pi^Y\downarrow^N_Y}}$ is a closure operator, we obtain that 
	\begin{equation}\label{exp:aux}
		\overline{X}_w\preceq_2 \overline{X}_w^{{\uparrow_\pi^Y\downarrow^N_Y}}=(T_Y)_w^{{\downarrow^N_Y}}\end{equation}
	
	Notice that, due to the solvability of the complete FRE and according to the first part of the proof, there exists a concept in $\CMcal{M}_{\pi N}(Y)$ of the form $(T_w^{\downarrow^N},(T_Y)_w)$. Hence, it holds that $T_w^{\downarrow^N}=(T_Y)_w^{{\downarrow^N_Y}}$ and, from inequality \eqref{exp:aux}, we deduce that 
	\begin{equation}\label{exp:aux2}
		\overline{X}_w\preceq_2(T_Y)_w^{{\downarrow^N_Y}}=T_w^{\downarrow^N}
	\end{equation}
	
	Now, as $^{{\uparrow_\pi\downarrow^N}}$ is a composition of isotone operators, $\overline{X}_w^{{\uparrow_\pi\downarrow^N}} \preceq_2 (T_w^{\downarrow^N})^{{\uparrow_\pi\downarrow^N}}$. According to the Galois connection properties of ${}^{\uparrow_\pi}$ and ${}^{\downarrow^N}$, the composition ${}^{\downarrow^N\uparrow_\pi\downarrow^N}$ equals ${}^{\downarrow^N}$. Therefore, we can assert that
	\begin{equation}
		\overline{X}_w^{{\uparrow_\pi\downarrow^N}} \preceq_2 (T_w^{\downarrow^N})^{{\uparrow_\pi\downarrow^N}}=T_w^{\downarrow^N}\end{equation}
	
	Suppose that $\overline{X}_w^{{\uparrow_\pi\downarrow^N}}\prec_2 T_w^{\downarrow^N}$ and we will obtain a contradiction. As $^{{\uparrow_\pi\downarrow^N}}$ is a closure operator, then
	$$\overline{X}_w\preceq_2 \overline{X}_w^{{\uparrow_\pi\downarrow^N}}\prec_2 T_w^{\downarrow^N}$$ 
	
	Moreover, from  expression \eqref{exp:aux2} we can assert  that $T_w^{\downarrow^N}=(T_Y)_w^{{\downarrow^N_Y}}$, and therefore
	\begin{equation}\label{exp:desig}\overline{X}_w\preceq_2 {X}_w^{{\uparrow_\pi\downarrow^N}}\prec_2 (T_Y)_w^{{\downarrow^N_Y}}\end{equation}
	
	Nevertheless, expression~\eqref{exp:desig} contradicts Proposition~\ref{prop:charact_predec}. In fact, since the pair $({X}_w^{{\uparrow_\pi\downarrow^N}},{X}_w^{{\uparrow_\pi}})$ is clearly a concept of $\CMcal{M}_{\pi N}(U)$ and $Y$ is a consistent set,
	there is a concept in $\CMcal{M}_{\pi N}(Y)$ whose extent is ${X}_w^{{\uparrow_\pi\downarrow^N}}$. In other words, there exists a concept in $\CMcal{M}_{\pi N}(Y)$ whose extent is strictly  contained between $\overline{X}_w$ and $(T_Y)_w^{{\downarrow^N_Y}}$.  As $\overline{X}$ is solution of the $Y$-reduced FRE, Proposition~\ref{prop:charact_predec} is contradicted.
	
	Consequently, $\overline{X_w}^{{\uparrow_\pi\downarrow^N}}= T_w^{\downarrow^N}$ for all $w\in W$. Applying $^{\uparrow_\pi}$ on both sides of the equality and taking into account Proposition~\ref{prop:charact_sol}, the following chain of equalities  holds  for all $w\in W$:  
	$$\overline{X_w}^{\uparrow_\pi}=\overline{X_w}^{{\uparrow_\pi\downarrow^N\uparrow_\pi}}= T_w^{\downarrow^N\uparrow_\pi}=T_w$$
	Therefore, $\overline{X}$ is solution of the complete FRE.
\end{proof}

Theorem~\ref{th:red-cons} states that reducing a FRE in a consistent set gives rise to an equivalent FRE, in the sense of having the same solution set. Consequently, reducing FRE via consistent sets enables to simplify their resolution without losing information.

In the context of simplifying the resolution of a FRE, we are interested in performing the greatest possible reduction. For this purpose, reducts are a type of consistent set of special interest, as they are minimal sets of attributes.

\begin{corollary}\label{cor:red-reduct}
	Let $R\odot_{\sigma}X=T$ be a solvable FRE and $Y$ a reduct of $(U,V,R,\sigma)$. We have that $Y$-reduced FRE of $R\odot_{\sigma}X=T$ is solvable. Moreover, $\overline{X}\in L_2^{V\times W}$ is a solution of the $Y$-reduced FRE if and only if it is a solution of the complete FRE.
\end{corollary}

Notice that, all reducts of a context do not necessarily have the same cardinality. Therefore, reducing a FRE via some reducts may result in a system with a shorter number of equations than reducing it in others, thus leading to a resolution with a lower computational cost. Consequently, reducts with the minimum number of attributes are the most appropriated consistent sets for the aim of reducing FRE.
	
Moreover, this reduction only depends on the associated context, so in certain systems where it is necessary dealing with several equations that share the coefficient matrix, the reduction must be performed just once. This is relevant when FRE are used in inference procedures like abduction. 

Next, a particular equation will be considered, reduced and solved in detail from the results introduced in this section. 

	\begin{example}\label{ex:reduction_consistent}
		Consider the sets $U=\{u_1,u_2,u_3,u_4,u_5\}$, $V=\{v_1,v_2,v_3,v_4,v_5\}$, $W=\{w\}$ and the multi-adjoint property-oriented frame
			\begin{equation*}([0,1]_8, \leq, \&^*_1, \swarrow^1_*,\nwarrow_1^*, \&_2^*, \swarrow^2_*, \nwarrow_2^*)\end{equation*}
			where $[0,1]_8$ denotes the regular partition of the unit interval in 8 pieces. The conjunction of the first adjoint triple is $\&^{*}_{1}\colon [0,1]_8\times[0,1]_8\rightarrow [0,1]_8$, defined as
\footnote{$\ulcorner\urcorner$ denotes the ceil function, that assigns to each value the first integer that is greater or equal than it. Analogously, $\llcorner\lrcorner$ denotes the floor function, that assigns to each value the first integer lower or equal than it.} $x\&^{*}_{1} y=\frac{\lceil 8x^2y\rceil}{8}$, for all $x,y\in  [0,1]_8$, and its corresponding residuated implications are $\swarrow_{*}^{1}\colon [0,1]_8\times[0,1]_8\rightarrow [0,1]_8$ and $\nwarrow^{*}_{1}\colon [0,1]_8\times[0,1]_8\rightarrow [0,1]_8$, that are defined, for all $x,y,z\in  [0,1]_8$, as
			$$z \swarrow_{*}^{1} y=\begin{cases}
				1 & \mbox{if $y=0$} \\
				\min\left\lbrace \frac{\lfloor 8\sqrt{z/y}\rfloor}{8},1\right\rbrace 
				& \mbox{otherwise}
			\end{cases}$$
			
			$$z \nwarrow^{*}_{1} x=\begin{cases}
				1 & \mbox{if $x=0$} \\
				\min\left\lbrace \frac{\lfloor 8z/x^2\rfloor}{8},1\right\rbrace 
				& \mbox{otherwise}
			\end{cases}$$
			
			{T}he second adjoint triple, $\&^{*}_{2}\colon [0,1]_8\times[0,1]_8\rightarrow [0,1]_8$ is defined by $x\&^{*}_{2} y=\frac{\lceil 8xy^2\rceil}{8}$, for all $x,y\in  [0,1]_8$, and its corresponding residuated implications are the mappings $\swarrow_{*}^{2}\colon [0,1]_8\times[0,1]_8\rightarrow [0,1]_8$  and $\nwarrow^{*}_{2}\colon [0,1]_8\times[0,1]_8\rightarrow [0,1]_8$, defined   for all $x,y,z\in  [0,1]_8$ as
			
			$$z \swarrow_{*}^{1} y=\begin{cases}
				1 & \mbox{if $y=0$} \\
				\min\left\lbrace \frac{\lfloor 8z/y^2\rfloor}{8},1\right\rbrace 
				& \mbox{otherwise}
			\end{cases}$$
			
			$$
			z \nwarrow^{*}_{1} x= \begin{cases}
				1 & \mbox{if $x=0$} \\
				\min\left\lbrace \frac{\lfloor 8\sqrt{z/x}\rfloor }{8},1\right\rbrace 
				& \mbox{otherwise}
			\end{cases}
			$$	
			
			Consider the FRE with sup-$\&_\sigma$-composition 
			\begin{equation}\label{exp:example_reduction_equation} 
				R\odot_\sigma X=T 
			\end{equation}
			where $\sigma\colon V\rightarrow\{1,2\}$ assigns $v_1, v_2, v_4$ to the first adjoint triple and $v_3, v_5$ to the second one,
			{\fontsize{10}{13}\selectfont $$R=\left( \begin{array}{ccccc}
					0.75 & 0.5 & 0 & 0.5 & 0.5 \\
					0.5 & 0.25 & 0.25 & 0.75 & 1 \\
					0.75 & 0.5 & 0.125 & 0 & 0.375 \\
					0.75 & 0.5 & 0 & 0.5 & 0.5 \\
					0.75 & 0.5 & 0.125 & 0 & 0.5 \\
				\end{array}\right), \quad T=
				\left( \begin{array}{c}
					0.25 \\
					0.5 \\
					0 \\
					0.25 \\
					0 \\
				\end{array}
				\right)$$}
			and $X\in [0,1]_8^{V\times W}$ is unknown. 
						
			We will check whether FRE~\eqref{exp:example_reduction_equation} is solvable by means of Proposition~\ref{prop:charact_solvable}, basing on its associated context $(U,V,R,\sigma)$. Specifically, we will see if $T =T^{\downarrow^N\uparrow_\pi}$, since $W$ is a singleton, that is, $T$ only has one column.  Indeed, making the corresponding computations, the following chain of equalities holds
			
				$$T^{\downarrow^N\uparrow_\pi} =\left( \begin{array}{c}
					0 \\
					0 \\
					0 \\
					0.875 \\
					0 \\
				\end{array}
				\right)\raisebox{30 pt}{\large$\uparrow_\pi$}= \left( \begin{array}{c}
					0.25 \\
					0.5 \\
					0 \\
					0.25 \\
					0 \\
				\end{array}
				\right)=T$$	
				
Following the procedures for calculating the reducts of a context detailed in~\cite{TFS:2020-acmr, Cornejo2017, ar:ins:2015}, we obtain that the context $(U,V,R,\sigma)$ admits two possible reducts: $Y_1=\{u_1, u_2, u_3\}$ and $Y_2=\{u_2, u_3, u_4\}$. Hence, by Corollary~\ref{cor:red-reduct}, we may consider two different ways of reducing FRE~\eqref{exp:example_reduction_equation} preserving its solution set.  Moreover, since the number of attributes in both reducts coincides, the reduced FRE corresponding to either $Y_1$ or $Y_2$ is optimal, in the sense that it contains the least possible number of equations. In what follows, we will show the details for reducing FRE~\eqref{exp:example_reduction_equation} with respect to $Y_1$, as the case related to $Y_2$ is completely analogous.

According to the reduct $Y_1=\{u_1, u_2, u_3\}$, it is sufficient preserving the three first equations of FRE~\eqref{exp:example_reduction_equation}, whilst  the last two equations can be removed. Hence, the reduced FRE in $Y_1$ is 
\begin{equation}\label{exp:reducSystem}R_{Y_1}\odot_\sigma X=T_{Y_1}\end{equation}
which corresponds to the system:

\begin{align*}
\notag 0.75 &\,\&_{1}\, x_1 &&\!\!\vee\!\! &  0.5&\,\&_{1} \, x_2&&\!\!\vee\!\! &  0&\,\&_{2} \, x_3 &&\!\!\vee\!\! &  0.5&\,\&_{1} \, x_4  && \!\!\vee\!\! & 0.5&\,\&_{2}\, x_5  &&\!\!=0.25\\
 0.5&\,\&_{1} \, x_1 &&\!\!\vee\!\! & 0.25&\,\&_{1} \, x_2&&\!\!\vee\!\! & 0.25&\,\&_{2} \, x_3&&\!\!\vee\!\! & 0.75&\,\&_{1} \, x_4&&\!\!\vee\!\! &  1&\,\&_{2}\, x_5   &&\!\!=0.5\\
\notag 0.75&\,\&_{1}\, x_1 &&\!\!\vee\!\! & 0.5&\,\&_{1} \, x_2&&\!\!\vee\!\! & 0.125&\,\&_{2} \, x_3&&\!\!\vee\!\! & 0&\,\&_{1} \, x_4&&\!\!\vee\!\! &  0.375&\,\&_{2}\, x_5 &&\!\!=0
 \end{align*}

Considering the context $(Y_1,V,R_{Y_1},\sigma_{\mid Y_1\times V})$ and making the corresponding computations, we obtain that its associated concept lattice contains 40 concepts, as shown in Figure~\ref{fig:complete_lattice_1}. Now, in order to apply Corollary~\ref{cor:charact_sol_set} to compute the solution set of FRE~\eqref{exp:reducSystem}, we first calculate the concept whose intent is 
	\[T_{Y_1}=\left( \begin{array}{c}
		0.25\\
		0.5\\
		0
	\end{array}
	\right)\]
	We can easily check that the extent of such concept is $(0,0,0,0.875,0)$, and the extent of its unique predecessor in Figure~\ref{fig:complete_lattice_1} is $(0,0,0,0.625,0)$, as depicted in Figure~\ref{fig:sublattice_example_1}. Therefore, by Corollary~\ref{cor:charact_sol_set}, the solution set of FRE~\eqref{exp:reducSystem} is given by
	\[\big((0,0,0,0.875,0)\big]\, \big\backslash\, \big((0,0,0,0.625,0)\big]\]
	that is the set
	\[\left\{X\in[0,1]_8^{V}\mid X\leq(0,0,0,0.875,0),X\not\leq(0,0,0,0.625,0)\right\}\]
	Consequently, we conclude that there are two solutions of the FRE~\eqref{exp:reducSystem}: 
\[X_1=\left( \begin{array}{c}
	0\\
	0\\
	0\\
	0.875\\
	0\\
\end{array}
\right)\qquad X_2=\left( \begin{array}{c}
	0\\
	0\\
	0\\
	0.75\\
	0\\
\end{array}
\right)\]
Equivalently, applying Corollary~\ref{cor:red-reduct}, $X_1$ and $X_2$ are the solutions of \linebreak FRE~\eqref{exp:example_reduction_equation}.\hfill\qedsymbol

\begin{figure}[h!]\caption{Concept lattice associated with FRE~\eqref{exp:reducSystem}}\label{fig:complete_lattice_1}
	\begin{minipage}{1\textwidth}
		\begin{center}
			\usetikzlibrary{shapes}
			\tikzstyle{place}=[circle,draw=black!75,fill=black!20,minimum width= 5 pt,align=center,minimum height=5]
			\begin{tikzpicture}[inner sep=2 mm,scale=1.5, every node/.style={scale=0.8}]		
				\node at (0,0) (c1) [place] {};
				
				\node at (-0.5,0.5) (c2) [place] {};
				\node at (0.5,0.5) (c3) [place] {};
				
				\node at (-1,1) (c4) [place] {};
				\node at (0,1) (c5) [place] {};
				\node at (1,1) (c6) [place] {};
				
				\node at (-0.5,1.5) (c7) [place] {};
				\node at (0.5,1.5) (c8) [place] {};
				
				\node at (-1.5,2) (c9) [place] {};
				\node at (0,2) (c10) [place] {};
				\node at (1,2) (c11) [place] {};
				
				\node at (-1,2.5) (c12) [place] {};
				\node at (0,2.5) (c13) [place] {};
				\node at (1,2.5) (c14) [place] {};
				
				\node at (-1.5,3) (c15) [place] {};
				\node at (-0.5,3) (c16) [place] {};
				\node at (0.5,3) (c17) [place] {};
				\node at (1.5,3) (c18) [place] {};
				
				\node at (-1,3.5) (c19) [place] {};
				\node at (0,3.5) (c20) [place] {};
				\node at (1,3.5) (c21) [place] {};
				
				\node at (-1.5,4) (c22) [place] {};
				\node at (-0.5,4) (c23) [place] {};
				\node at (1,4) (c24) [place] {};
				
				\node at (-2,4.5) (c25) [place] {};
				\node at (-1,4.5) (c26) [place] {};
				\node at (0.375,4.5) (c27) [place] {};
				\node at (1.5,4.5) (c28) [place] {};
				
				\node at (-1.5,5) (c29) [place] {};
				\node at (-0.25,5) (c30) [place] {};
				\node at (1,5) (c31) [place] {};
				
				\node at (-1.5,5.5) (c32) [place] {};
				\node at (-0.25,5.5) (c33) [place] {};
				\node at (1,5.5) (c34)[place] {};
				
				\node at (-1.5,6) (c35) [place] {};
				\node at (-0.25,6) (c36) [place] {};
				\node at (1,6) (c37)[place] {};
				
				\node at (-0.875,6.5) (c38) [place] {};
				\node at (0.375,6.5) (c39) [place] {};
				
				\node at (-0.25,7) (c40) [place] {};	
				
				\draw [-] (c1) -- (c2) -- (c1) -- (c3);
				\draw [-] (c2) -- (c4) -- (c2) -- (c5); \draw [-] (c3) -- (c4) -- (c3) -- (c6);
				\draw [-] (c4) -- (c9) -- (c4) -- (c7) -- (c5); \draw [-] (c6) -- (c7) -- (c6) -- (c8);
				\draw [-] (c8) -- (c11) -- (c8) -- (c10) -- (c7) -- (c12);
				\draw [-] (c9) -- (c15) -- (c9) -- (c12);
				\draw [-] (c10) -- (c13) -- (c11) -- (c14);
				\draw [-] (c12) -- (c19) -- (c12) -- (c16) -- (c13) -- (c17) -- (c14) -- (c18);
				\draw [-] (c15) -- (c19);
				\draw [-] (c16) -- (c23) -- (c16) -- (c20) -- (c17) -- (c21) -- (c18);
				\draw [-] (c19) -- (c22) -- (c19) -- (c23);
				\draw [-] (c20) -- (c27) -- (c20) -- (c24);
				\draw [-] (c21) -- (c24) -- (c28);
				\draw [-] (c22) -- (c25) -- (c22) -- (c26) -- (c23) -- (c27);
				\draw [-] (c25) -- (c29) -- (c26) -- (c30) -- (c27) -- (c31) -- (c28);
				\draw [-] (c29) -- (c32) -- (c30) -- (c33) -- (c31) -- (c34);
				\draw [-] (c32) -- (c35) -- (c33) -- (c36) -- (c34) -- (c37);
				\draw [-] (c35) -- (c38) -- (c36) -- (c39) -- (c37);
				\draw [-] (c38) -- (c40) -- (c39);
			\end{tikzpicture}
		\end{center}
	\end{minipage}
\end{figure}
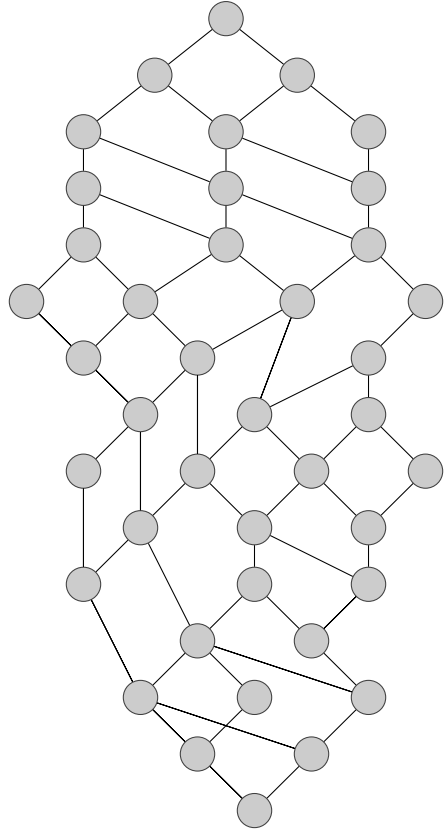
		\begin{figure}[h!]
		\caption{{Sublattice} associated with FRE~\eqref{exp:reducSystem}}\label{fig:sublattice_example_1}
				\begin{minipage}{1\textwidth}
					\begin{center}
						\hspace{20 pt}
						\includegraphics*[scale=0.5]{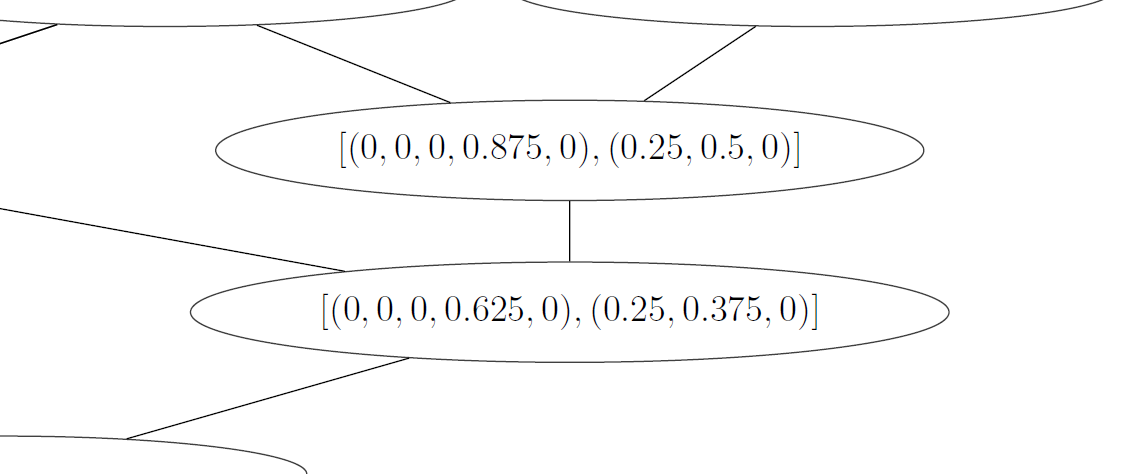}
					\end{center}
				\end{minipage}
			\end{figure} 

\end{example}
Due to Theorem~\ref{th:red-cons}, given a solvable FRE and a consistent set of its associated context, we can remove unnecessary equations and compute its whole solution set from the resulting reduced FRE. It is natural to wonder what happens if the considered attributes set in the reduction process is non-consistent. As shown in the example below, in general, the solution set of the complete and the reduced FRE do not coincide.

	\begin{example}
		Given FRE~\eqref{exp:example_reduction_equation}  introduced in Example~\ref{ex:reduction_consistent} and the attributes set $Y_3=\{u_3,u_4\}$, which is clearly a non-consistent set of $(U,V,R,\sigma)$, since it is strictly contained in the reduct $Y_2$.
		
		\begin{figure}[h!]\caption{Sublattice associated with FRE $R_{Y_3}\odot_\sigma X=T_{Y_3}$}\label{fig:sublattice_example_2}
			\begin{minipage}{1\textwidth}
				\begin{center}
					\hspace{20 pt}
					\includegraphics*[scale=0.35]{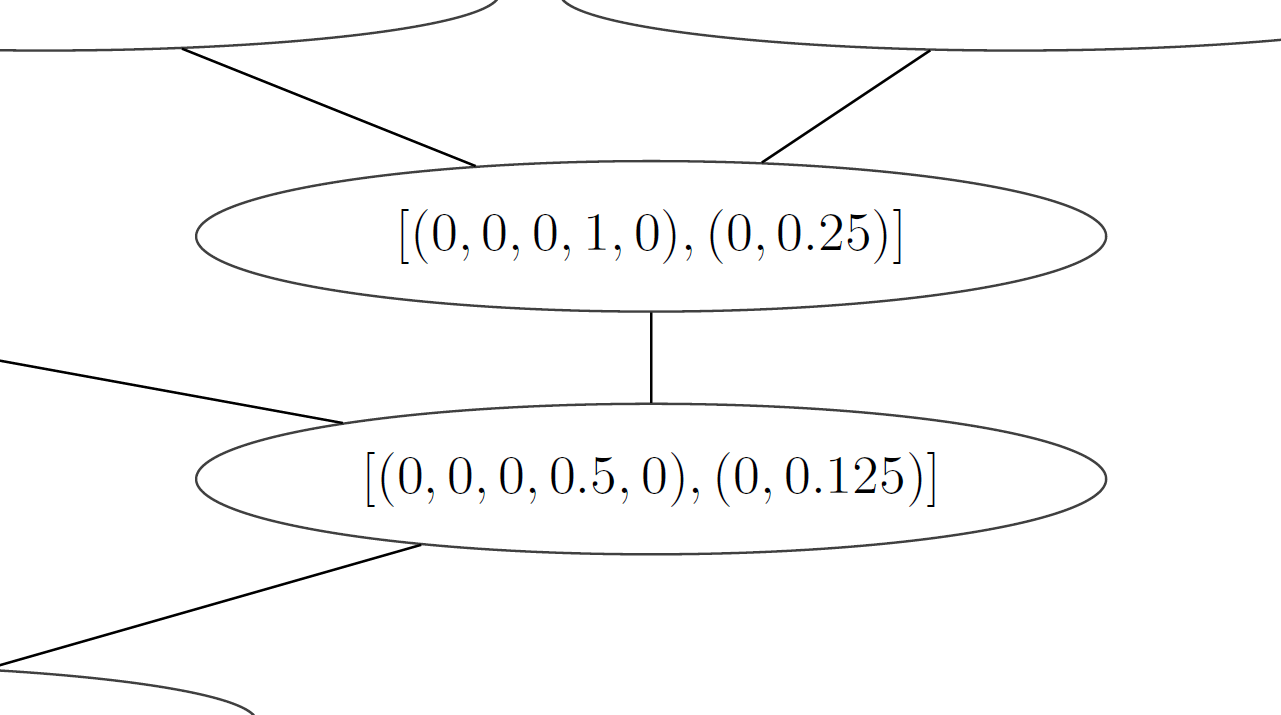}
				\end{center}
			\end{minipage}
		\end{figure}
		First of all, from the concept lattice associated with $R_{Y_3}\odot_\sigma X=T_{Y_3}$, we fix our attention to the sublattice depicted in Figure~\ref{fig:sublattice_example_2},  which contains the concept with the intent with values $0.25$ and $0$ for attributes $u_3$ and $u_4$, respectively,   and its precedent concepts (only one in this case). According to Corollary~\ref{cor:charact_sol_set}, the whole solution set is
			\[\big((0,0,0,1,0)\big]\, \big\backslash\, \big((0,0,0,0.5,0)\big]\]
			that is the set
			\[\left\{X\in[0,1]_8^{V}\mid X\leq(0,0,0,1,0),X\not\leq(0,0,0,0.5,0)\right\}\] obtaining the following four solutions:
			$$
			X_1=\left( \begin{array}{c}
				0\\
				0\\
				0\\
				1\\
				0\\
			\end{array}
			\right)\quad X_2=\left( \begin{array}{c}
				0\\
				0\\
				0\\
				0.875\\
				0\\
			\end{array}
			\right) \quad X_3=\left( \begin{array}{c}
				0\\
				0\\
				0\\
				0.75\\
				0\\
			\end{array}
			\right)\quad X_4=\left( \begin{array}{c}
				0\\
				0\\
				0\\
				0.625\\
				0\\
			\end{array}
			\right)$$
			
			Clearly, $X_1$ and $X_4$ are solutions of the reduced equation but they are not solutions of the complete FRE~\eqref{exp:example_reduction_equation}. This happens because $Y_3$ is not a consistent subset of attributes of the associated context and, as a consequence,  information is lost when only considering equations three and \linebreak four. \hfill\qedsymbol
	\end{example}
\section{Approximation of unsolvable multi-adjoint FRE}\label{sec:approximation}
The results developed in Section~\ref{sec:reduction} deal with the reduction of solvable FRE. In what follows, we will show how the reduction theory presented in this paper can be used to reduce an unsolvable FRE, i.e. to eliminate equations, so that it becomes solvable. In this case, we may reason that the eliminated equations correspond to incoherences in the data. Notice that, as it has been shown in Section~\ref{sec:reduction}, a MARE cannot be reduced in an arbitrary way if we want to preserve the information. Consistent sets and reducts will also play a key role in the reduction process in order to ensure that no information is being lost when approximating a MARE.

Given an unsolvable FRE
\begin{equation}\label{eq:introapr}
R\odot_\sigma X=T
\end{equation}
associated with the context $(U,V,R,\sigma)$ and suppose that there exists a set $Y\subseteq U$ such that the $Y$-reduced FRE
\[R_Y\odot_{\sigma}X=T_Y\]
is solvable. In that case, we may assert that the existing incoherences in FRE~\eqref{eq:introapr} have been eliminated when carrying out the reduction. Nevertheless, we might have removed more than needed. For example, the $\varnothing$-reduced FRE is clearly solvable, since all equations of FRE~\eqref{eq:introapr} are eliminated, but then all information contained in FRE~\eqref{eq:introapr} has been deleted.

We will say that an equation is ``problematic'' if removing it from an unsolvable MARE results in a solvable one, what means that unsolvability was induced by this equation. However, it only makes sense removing a problematic equation if it is not associated with an attribute in a reduct, i.e., if it contains redundant information in the coefficient matrix.
	
Therefore, we are interested in removing the equations that are redundant and problematic in MARE~\eqref{eq:introapr}, whilst preserving the rest of equations. To this aim, following the approach provided in Theorem~\ref{th:red-cons}, we will demand the ground set $Y$ to be a consistent set of $(U,V,R,\sigma)$. Furthermore, in order to also remove redundant equations (in their left side), we will require $Y$ to be a reduct of $(U,V,R,\sigma)$.
\begin{definition}\label{def:feasible_reduct}
	Let $R\odot_{\sigma}X=T$ be an unsolvable {FRE} and $(U,V,R,\sigma)$ its associated context. A reduct $Y$ {of} $(U,V,R,\sigma)$ is \emph{feasible} if the $Y$-reduced FRE $R_Y\odot_{\sigma}X=T_Y$ is solvable. 
\end{definition}

Notice that the  notion of feasible  consistent set $Y$  could be similarly defined. However, we have considered only feasible reducts due to the minimal character of them and the following property. 

\begin{proposition}\label{prop:feasible-red}
	Let $R\odot_{\sigma}X=T$ be an unsolvable {FRE} and $(U,V,R,\sigma)$ its associated context. If $Y\subseteq U$ is a consistent set  such that the $Y$-reduced equation $R_Y\odot_{\sigma}X=T_Y$ is solvable, then there exists a feasible reduct $Y'\subseteq Y$.
\end{proposition}
\begin{proof}

First of all, we take into account that, since $U$ is finite, clearly every consistent set contains, at least, a reduct. Hence, given the consistent set $Y\subseteq U$, we consider a reduct  $Y'\subseteq Y$ of $(U,V,R,\sigma)$.
 
Now, as the $Y$-reduced FRE $R_Y\odot_{\sigma}X=T_Y$ is solvable and $Y'\subseteq Y$ is a reduct of $(U,V,R,\sigma)$ and thus of $(Y,V,R_Y,\sigma_{\mid Y\times V})$, applying Corollary~\ref{cor:red-reduct} we conclude that $R_{Y'}\odot_{\sigma}X=T_{Y'}$ is solvable. In other words, $Y'$ is a feasible reduct of $(U,V,R,\sigma)$.

\end{proof}

As discussed at the beginning of Section~\ref{sec:approximation}, if $Y\subseteq U$ is a feasible reduct of an unsolvable FRE
\begin{equation}\label{eq:thapr}
R\odot_\sigma X=T
\end{equation}
then we can deduce that the problematic equations of FRE~\eqref{eq:thapr} are not present in the $Y$-reduced FRE
\begin{equation}\label{eqred:thapr}
{R_Y}\odot_{\sigma}X=T_Y
\end{equation}
It is natural to wonder what is the incoherence of the missing equations due to. For example, suppose that we modify the elements in the independent term of FRE~\eqref{eq:thapr} that are associated with the eliminated equations, that is, we conveniently change the independent terms related to $U\setminus Y$, to obtain a solvable FRE
\begin{equation}\label{eq2:thapr}
R\odot_\sigma X=T^*
\end{equation}
In that case, we say that FRE~\eqref{eq2:thapr} is a \emph{solvable approximation} of FRE~\eqref{eq:thapr}.

The following theorem shows that there always exists a solvable approximation of an unsolvable FRE if there exists at least a feasible reduct. Indeed, there exists a solvable approximation for each feasible reduct. Furthermore, basing on the fact that FRE~\eqref{eqred:thapr} is also a $Y$-reduced version of FRE~\eqref{eq2:thapr}, the term $T^*$ can be defined as $T_Y^{\downarrow_Y^N\uparrow_\pi}$.

\begin{theorem}\label{th:approximation}
	Let $R\odot_{\sigma}X=T$ be an unsolvable {FRE} and $Y$ a feasible reduct of its associated context.  {There} exists $T^*\in L_1^{U\times W}$  {such}   that $R\odot_{\sigma}X=T^*$ is solvable and ${R_Y}\odot_{\sigma}X=T_Y$ is a $Y$-reduced {FRE} {of} $R\odot_{\sigma}X=T^*$. Additionally, $T^*(u,w)=(T_Y)_w^{\downarrow_Y^N\uparrow_\pi}(u)$, for all $(u,w)\in U\times W$.
\end{theorem}
\begin{proof}

 Let $T^*\in L_1^{U\times W}$ be the relation given, for all $(u,w)\in U\times W$, by 
\[T^*(u,w)=(T_Y)_w^{\downarrow_Y^N\uparrow_\pi}(u)\]
Firstly, we will see that the FRE $R\odot_{\sigma}X=T^*$ is solvable. Equivalently, according to Proposition~\ref{prop:charact_solvable}, we will prove that $T^*_w\in \CMcal{I}(\CMcal{M}_{\pi N}(U))$ for each $w\in W$.
	
Since $Y$ is a feasible set, the $Y$-reduced FRE ${R_Y}\odot_{\sigma}X=T_Y$ is solvable. Therefore, by Proposition~\ref{prop:charact_solvable}, $(T_Y)_w\in \CMcal{I}(\CMcal{M}_{\pi N}(Y))$ for each $w\in W$, and thus $(T_Y)_w^{\downarrow_Y^N}\in \CMcal{E}(\CMcal{M}_{\pi N}(Y))$. Furthermore, as $Y$ is a reduct, 
\[\CMcal{M}_{\pi N}(U)\cong_\CMcal{E} \CMcal{M}_{\pi N}(Y)\]
from which $(T_Y)_w^{\downarrow_Y^N}\in \CMcal{E}(\CMcal{M}_{\pi N}(U))$. Hence, $T^*_w=(T_Y)_w^{\downarrow_Y^N\uparrow_\pi}\in \CMcal{I}(\CMcal{M}_{\pi N}(U))$.
	
Now, we will see that ${R_Y}\odot_{\sigma}X=T_Y$ is a $Y$-reduced FRE of $R\odot_{\sigma}X=T^*$. To this aim, we will prove that $(T_Y)^*_w(y)=(T_Y)_w(y)$ for each $y\in Y$, or equivalently, $(T_Y)_w^{\downarrow_Y^N\uparrow_\pi}(y)=(T_Y)_w(y)$ for each $y\in Y$.
		
	As a consequence of Remark~\ref{rem:possibility}, 
	\[(T_Y)_w^{\downarrow_Y^N\uparrow_\pi^Y}(y)=(T_Y)_w^{\downarrow_Y^N\uparrow_\pi}(y)\]
	for each $y\in Y$. Moreover, since $(T_Y)_w\in \CMcal{I}(\CMcal{M}_{\pi N}(Y))$, we can assert that $(T_Y)_w^{\downarrow_Y^N\uparrow_\pi^Y}=(T_Y)_w$. Therefore, the following chain of equalities holds, for each $y\in Y$:
	\[(T_Y)_w^{\downarrow_Y^N\uparrow_\pi}(y)=(T_Y)_w^{\downarrow_Y^N\uparrow_\pi^Y}(y)=(T_Y)_w(y)\]
We conclude that ${R_Y}\odot_{\sigma}X=T_Y$ is a $Y$-reduced FRE of $R\odot_{\sigma}X=T^*$.
\end{proof}	
	
Theorem~\ref{th:approximation} leads to a new way of dealing with unsolvable FRE, as every feasible reduct is related to a solvable approximation. The possibility of providing different approximations of an unsolvable FRE is specially interesting when the underlying data (giving rise to the FRE) present incoherences or are obtained from imprecise measurements. Approximations serve as different alternatives to mend the incoherences or the measurement errors.
	
Notice that the procedure of computing consistent sets/reducts does not depend on $T$, therefore, this procedure can also be applied to solving a set of equations with the same matrix $R$.
Moreover, it is expected not all the reducts are necessarily feasible, that is, only some of them might provide an approximation to the equation.

The following example illustrates the computation of approximations of an unsolvable FRE as well as the inference process to detect what might be causing its unsolvability and how could it be restored.
 
\begin{example}\label{ex:approximation}
Given the FRE with sup-$\&_\sigma$-composition 
\begin{equation}\label{exp:example_approx_equation} 
	R\odot_\sigma X=T
\end{equation}
where  
$$T=
\left( \begin{array}{c}
	0.5\\0.875\\0.375\\0.625\\0.125
\end{array}
\right)$$
and the multi-adjoint frame, the context and $R$ are the same as in Example~\ref{ex:reduction_consistent}. Therefore, the context and the concept lattice will remain the same. Applying Proposition~\ref{prop:charact_solvable}, the solvability of FRE~\eqref{exp:example_approx_equation} is equivalent to the satisfiability of the equality $T =T^{\downarrow^N\uparrow_\pi}$. Nevertheless, making the corresponding computations:

\[T^{\downarrow^N\uparrow_\pi}=\left( \begin{array}{c}
	0.125 \\
	0.5 \\
	1 \\
	1 \\
	0.5 \\
\end{array}
\right)\raisebox{30 pt}{\large$\uparrow_\pi$}=\left( \begin{array}{c}
	0.25 \\
	0.625 \\
	0.125 \\
	0.25 \\
	0.125 \\
\end{array}
\right)\neq T\]

Consequently, FRE~\eqref{exp:example_approx_equation} is unsolvable. According to Theorem~\ref{th:approximation}, we will find now the (solvable) approximations of FRE~\eqref{exp:example_approx_equation}.
From Example~\ref{ex:reduction_consistent}, the context associated with FRE~\eqref{exp:example_approx_equation} admits two possible reducts, $Y_1=\{u_1,u_2,u_3\}$ and $Y_2=\{u_2,u_3,u_4\}$, so two different reduced FRE can be considered. In order to see if $Y_1$ and $Y_2$ are feasible reducts of  FRE~\eqref{exp:example_approx_equation}, we will study the solvability of their corresponding reduced FRE.

Concerning the reduct $Y_1$, we obtain that 
\[T_{Y_1}^{\downarrow^N_{Y_1}\uparrow_\pi^{Y_1}}=\left( \begin{array}{c}
	0.625 \\
	1 \\
	1 \\
	1 \\
	0.875 \\
\end{array}
\right)\raisebox{30 pt}{\large$\uparrow_\pi^{Y_1}$}=\left( \begin{array}{c}
	0.5\\0.875\\0.375\end{array}
\right)= T_{Y_1}\]
 By Proposition~\ref{prop:charact_solvable}, $R_{Y_1}\odot_\sigma X=T_{Y_1}$ is solvable, and thus $Y_1$ is a feasible reduct. According to Theorem~\ref{th:approximation}, this fact leads to an approximation of FRE~\eqref{exp:example_approx_equation}, whose independent term is given by
\[T^{*}=T^{\downarrow^N_{Y_1}\uparrow_\pi}=
\left( \begin{array}{c}
	0.5 \\
	0.875 \\
	0.375 \\
	0.5 \\
	0.5 \\\end{array}\right)\]

Now,  the solution set of this FRE can be computed using Corollary~\ref{cor:charact_sol_set}, as we show next.  Notice that the concept lattice associated with FRE~\eqref{exp:example_approx_equation} is the same as the one obtained in Example~\ref{ex:reduction_consistent}. The part of the concept lattice in which we are interested in is displayed in Figure~\ref{fig:sublattice_example_3}.

\begin{figure}[h!]\caption{Part of the concept lattice associated with  FRE~\eqref{exp:example_approx_equation}}\label{fig:sublattice_example_3}
	\begin{minipage}{1\textwidth}
		\begin{center}
			\hspace{20 pt}
			\includegraphics*[scale=0.64]{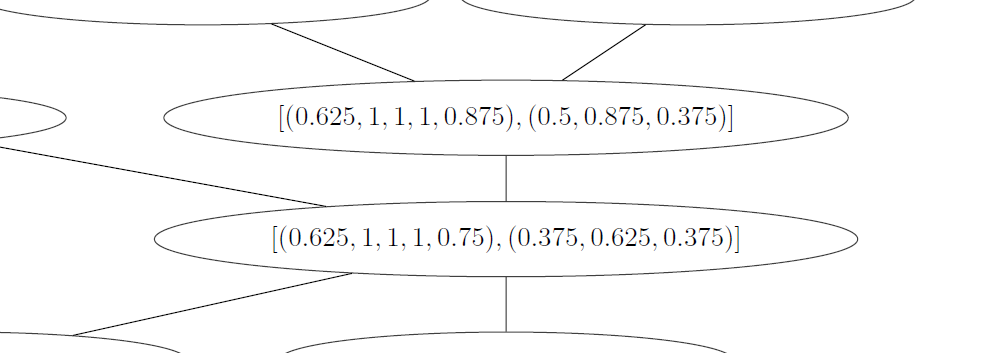}
		\end{center}
	\end{minipage}
\end{figure}

 In this case, we will omit the computation of the solution set due to its large number of elements. There exist 4734 solutions of the approximated equation, being the maximum and minimum ones
	$$X_1=\left( \begin{array}{c}
		0.625\\1\\1\\1\\0.875\end{array}
	\right) \qquad X_{4734}=\left( \begin{array}{c}
		0\\0\\0\\0\\0.875\end{array}
	\right)$$
	
 {With regard to the reduct $Y_2=\{u_2,u_3,u_4\}$, we obtain that}

\[T_{Y_2}^{\downarrow^N_{Y_2}\uparrow_\pi^{Y_2}}=\left( \begin{array}{c}
	0.625 \\
	1 \\
	1 \\
	1 \\
	0.875 \\
\end{array}
\right)\raisebox{30 pt}{\large$\uparrow_\pi^{Y_2}$}=\left( \begin{array}{c}
	0.875\\0.375\\0.5\end{array}
\right)\neq T_{Y_2}\]
 Consequently, by Proposition~\ref{prop:charact_solvable}, we conclude that the $Y_2$-reduced FRE $R_{Y_2}\odot_\sigma X=T_{Y_2}$ is not solvable, and therefore the unique approximation of FRE~\eqref{exp:example_approx_equation} is the one related to $Y_1$.
\qed
\end{example}

\begin{remark}\label{rem:approximation}
Example~\ref{ex:approximation} admits an interpretation of what  causes the unsolvability of FRE~\eqref{exp:example_approx_equation}. As $Y_1=\{u_1,u_2,u_3\}$ is its unique feasible reduct, there exists some kind of contradiction between the first three equations (related to $\{u_1,u_2,u_3\}$) and the last two equations (related to $\{u_4,u_5\}$) of FRE~\eqref{exp:example_approx_equation}. Indeed, if we look at the FRE~\eqref{exp:example_approx_equation} row by row, a hypothetical solution $(x_1,x_2,x_3,x_4,x_5)$ should satisfy the equations:
	
	\begin{align*}\begin{pmatrix}
			0.75 & 0.5 & 0 & 0.5 &0.5
		\end{pmatrix}\odot_\sigma \begin{pmatrix}
			x_1 \\ x_2 \\ x_3 \\ x_4 \\ x_5
		\end{pmatrix}&=\begin{pmatrix}
			0.5
		\end{pmatrix}\hspace{3em} \mathrm{(related\ to\ }u_1\in Y_1\mathrm{)}\\
		\begin{pmatrix}
			0.75 & 0.5 & 0 & 0.5 &0.5
		\end{pmatrix}\odot_\sigma \begin{pmatrix}
			x_1 \\ x_2 \\ x_3 \\ x_4 \\ x_5
		\end{pmatrix}&=\begin{pmatrix}
			0.625
		\end{pmatrix}\hspace{3em} \mathrm{(related\ to\ }u_4\notin Y_1\mathrm{)}
	\end{align*}
	
	Clearly, the two previous equalities are incompatible, from which the unsolvability of FRE~\eqref{exp:example_approx_equation} arises. A similar conclusion can be reached with respect to the equations related to $\{u_2,u_3\}\subseteq Y_1$ and the one related to $u_5\not\in Y_1$.
	
According to the form of $T^*$ in Example~\ref{ex:approximation}, replacing $T(u_4)=0.625$ by $T^*(u_4)=0.5$ and $T(u_5)=0.125$ by $T^*(u_5)=0.5$ we eradicate the incompatibilities of FRE~\eqref{exp:example_approx_equation}, which turns out to be solvable with the new independent term. From an applicative perspective, if the values $T(u_4)$ and $T(u_5)$ have been obtained by measurements, we may assert that the measurement process of $u_5$ is notably inaccurate and should be checked, while the measure of $u_4$ is slightly imprecise.
	
Lastly, since the reduct $Y_2=\{u_2, u_3, u_4\}$ is not feasible, we can conclude that FRE~\eqref{exp:example_approx_equation} will be unsolvable whenever the values of $u_1$ and $u_5$ remain the same, no matter the values associated with $u_2$, $u_3$ and $u_4$
	
In conclusion, in case that the different conditions required in a FRE are incoherent, reduct approximation enables us in some particular situations to take into account only necessary variables, providing an approximation for the whole equation if the reduced equation is solvable.  
\end{remark}

	The contribution of this paper has a certain connection with the works presented in~\cite{CornejoFRE2017}, where the authors develop methods to approximate an unsolvable FRE from a different perspective. The foundations of these works  consists of replacing every column of the independent term  $T_w$ (interpreted as a mapping) of the unsolvable FRE by the intent of a concept, which is close to $T_w$. In~\cite{CornejoFRE2017}, these intent were selected following two philosophies, providing the pessimistic and optimistic approximations.
	
Coming back to Example~\ref{ex:approximation} and following the pessimistic procedure presented in~\cite{CornejoFRE2017}, the independent term 
	\[T=\left( \begin{array}{c}
		0.5\\0.875\\0.375\\0.625\\0.125
	\end{array}
	\right)\]
	would be replaced by 
	\[T_w^{\downarrow^N\uparrow_\pi}=\left( \begin{array}{c}
		0.25 \\
		0.625 \\
		0.125 \\
		0.25 \\
		0.125 \\
	\end{array}
	\right)\]
	The resulting FRE $R\odot_\sigma X= T_w^{\downarrow^N\uparrow_\pi}$ is solvable and therefore it can be seen as a solvable approximation of FRE~\eqref{exp:example_approx_equation}. As shown in Example~\ref{ex:approximation}, our reduction method revealed another approximation of FRE~\eqref{exp:example_approx_equation}, whose independent term is given by
	\[T^*=\left( \begin{array}{c}
		0.5 \\
		0.875 \\
		0.375 \\
		0.5 \\
		0.5 \\\end{array}\right)\]
	Comparing both approximations, we can highlight the following features:

	\begin{itemize}
	\item The method presented in~\cite{CornejoFRE2017} may result in a modification of all elements in $T$, whilst the approximation method based on feasible reducts only modify some equations. In particular, it leaves untouched the equations related to a feasible reduct.
	\item Since $T_w^{\downarrow^N\uparrow_\pi}\preceq T_w$, for each $w\in W$, the procedure shown in~\cite{CornejoFRE2017} only modifies the elements in $T$ by reducing their values (pessimistic) or increasing their values (optimistic). On the contrary, there are no such requirements in the reduction method. For example, 
	\[T^*(u_5)=0.5\not\preceq 0.125=T(u_5)\]
	\end{itemize}

\section{Reduction and approximation in a dual multi-adjoint FRE}\label{sec:dual}
A dual equation can be developed if we consider the same composition $\odot_\sigma$, and the  equality $R\odot_\sigma S=T$, where $R$ is the unknown fuzzy relation. This section will study and adapt the definitions and results  given above to the equation  $X\odot_\sigma S=T$.

Let $U,V,W$ be finite sets and $(L_1, L_2, P,\&_1,\dots, \&_n)$ a fixed multi-adjoint object-oriented frame~\cite{ins-medina}. Consider $X\odot_\sigma S=T$ a FRE, where $S\in P^{V\times W}$, $T\in L_2^{U\times W}$ and $X\in L_1^{U\times V}$, being $X$ unknown, and let $(V,W,S,\sigma)$ be its associated multi-adjoint context.

This equation admits dual results to the ones provided in the preliminaries section, which can be found {in}~\cite{dm:mare, dm:ins2014}. The results proved in Section~\ref{sec:reduction} and Section~\ref{sec:approximation}, that reduce and approximate equations of the form $R\odot_\sigma X=T$ can be adapted to this equation too. The main difference {is} in the notion of $Y$-reduced equation, which is given using object reduction instead of attribute reduction. Now, we will introduce the definition of reduced FRE with respect to an $\CMcal{I}$-reduct.
\begin{definition}\label{def:reduc_FRE_dual}
	Let $Y\subseteq W$ and consider the relations $S_Y=S_{\mid V\times Y}$ and $T_Y=T_{\mid U \times Y}$. The  multi-adjoint FRE $X\odot_\sigma S_Y=T_Y$ is called \emph{$Y$-reduced FRE of $X\odot_\sigma S=T$}.
\end{definition}
The following result is equivalent to Theorem~\ref{th:red-cons}, proving that the solutions of the  equations reduced by consistent sets are solutions of the  solvable complete one and vice versa.
\begin{theorem}\label{th:red_cons_dual}
	Let $X\odot_{\sigma}S=T$ be a solvable {FRE} and $Y$ an $\CMcal{I}$-consistent set of $(V,W,S,\sigma)$. The $Y$-reduced FRE of $X\odot_{\sigma}S=T$ is solvable. Moreover, $\overline{X}\in L_1^{U\times V}$ is a solution of the $Y$-reduced FRE {if and only if} it is a solution of the complete FRE. 
\end{theorem}
\begin{proof}
	Analogous to Theorem~\ref{th:red-cons} with just considering the multi-adjoint object-oriented concept lattice and its associated definitions.
\end{proof}

The approximation procedure introduced in the previous section can also be replicated. First of all, the notion of \emph{feasible reduct} is given in this framework.

\begin{definition}\label{def:feasible:reduct:dual}
	Let $X\odot_{\sigma}S=T$ be an unsolvable FRE and $Y$ an $\CMcal{I}$-reduct of $(V,W,S,\sigma)$. We will call this $\CMcal{I}$-reduct \emph{feasible} if the reduced equation $X\odot_{\sigma}S_Y=T_Y$ is solvable.
\end{definition} 

The following result is also dual to Theorem~\ref{th:approximation} for the particular FRE considered in this section. Unsolvable equations can be approximated by means of a reduced equation that preserves the essential information of the complete one.

\begin{theorem}\label{th:approximation_dual}
	Let $X\odot_{\sigma}S=T$ be an unsolvable FRE and $Y$ a feasible $\CMcal{I}$-reduct of the associated context. Then, there exists $T^*\in L_2^{U\times V}$ satisfying that $X\odot_{\sigma}S=T^*$ is solvable and ${X}\odot_{\sigma}S_Y=T_Y$ is an $Y$-reduced FRE of $X\odot_{\sigma}S=T^*$. Additionally, $T^*(u,w)=(T_Y)_w^{\uparrow_Y^N\downarrow^\pi}(u)$, for all $(u,w)\in U\times W$.
\end{theorem}
\begin{proof}
	Analogous to Theorem~\ref{th:approximation} with just considering the multi-adjoint object-oriented concept lattice and its associated definitions.
\end{proof}

Provided the hypothesis of Theorem~\ref{th:red_cons_dual}, we say that $X\odot_{\sigma}S=T^*$ is an \emph{approximation in the $\CMcal{I}$-reduct $Y$ of the FRE $X\odot_{\sigma}S=T$}.

\section{Reducing and approximating a FRE with max-min composition} \label{sec:classical_FRE}
All the results that have been introduced in this paper have been developed under a multi-adjoint frame, which has been chosen because of its flexibility. However, the presented results are also applicable in the classical case of a max-min FRE, with just considering the unit interval and  the Gödel t-norm in the multi-adjoint frame, that is, considering the Heyting algebra $([0,1],\adjoint_\G)$.  We will denote as $\odot_\G$ the composition operator in which all variables are assigned to the Gödel t-norm, that is, the max-min composition.

The first result that can be proved in this environment is the one related to the reduction of FRE. 

\begin{theorem}\label{th:red_cons_maxmin}
	Let $R\odot_\G X=T$ be a solvable {FRE} and $Y$ a consistent set of $(U,V,R)$. The $Y$-reduced FRE of $R\odot_\G X=T$ is solvable. Moreover, $\overline{X}\in L_2^{V\times W}$ is a solution of the $Y$-reduced FRE if and only if it is a solution of the complete FRE. 
\end{theorem}
\begin{proof}
	Particular case of Theorem~\ref{th:red-cons}.
\end{proof}

In this case, it makes no sense considering a dual FRE, as the composition operator $\odot_\G$ is commutative.

The definition of feasible reduct can be naturally extended from Definition~\ref{def:feasible_reduct}, as it is not affected by considering a residuated lattice.

\begin{definition}\label{def:feasible_reduct_maxmin}
	Let $R\odot_\G X=T$ be an unsolvable {FRE} and $(U,V,R)$ its associated context. A reduct $Y$ {of} $(U,V,R)$ is \emph{feasible} if the $Y$-reduced FRE $R_Y\odot_\G X=T_Y$ is solvable. 
\end{definition}

The approximation of unsolvable FRE can also be naturally particularized to the max-min composition.

\begin{theorem}\label{th:approximation_maxmin}
	Let $R\odot_\G X=T$ be an unsolvable {FRE} and $Y$ a feasible reduct of its associated context.  There exists $T^*\in L_1^{U\times W}$  {such}   that $R\odot_\G X=T^*$ is solvable and ${R_Y}\odot_\G X=T_Y$ is a $Y$-reduced {FRE} {of} $R\odot_\G X=T^*$. Additionally, $T^*(u,w)=(T_Y)_w^{\downarrow_Y^N\uparrow_\pi}(u)$, for all $(u,w)\in U\times W$.
\end{theorem}
\begin{proof}
	Particular case of Theorem~\ref{th:approximation}.
\end{proof}

Finally, an example will illustrate how this results are applied to the case of FRE defined with the max-min composition.

\begin{example}
	Given the sets $U=\{u_1,u_2,u_3,u_4,u_5\}$, $V=\{v_1,v_2,v_3,v_4,v_5\}$, $W=\{w\}$, and   the granular interval $[0,1]_8$, we  consider the following FRE 
	\begin{equation}\label{exp:eq_max_min}
		R\odot_\text{G} X=T 
	\end{equation}
	 where
	{\fontsize{10}{13}\selectfont $$R=\left( \begin{array}{ccccc}
			0.5 & 0.25 & 0.75 & 0.675 & 0.25 \\
			0.25 & 0.5 & 0.75 & 0.5 & 0.375 \\
			0.125 & 0.5 & 0.75 & 0.5 & 0.5 \\
			0.25 & 0.5 & 0.5 & 0.5 & 0.375 \\
			0.5 & 0.25 & 0.75 & 0.5 & 0.25 \\
		\end{array}\right), \quad T=
		\left( \begin{array}{c}
			0.5 \\
			0.375 \\
			0.375 \\
			0.375 \\
			0.5 \\
		\end{array}
		\right)$$}
	and $X\in [0,1]_8^{V\times W}$ is unknown. 
	
	First of all, it can be checked that FRE~\eqref{exp:eq_max_min} is solvable, as the following equality holds, considering the associated formal context to FRE~\eqref{exp:eq_max_min}, that is, $(U,V,R)$
	
	$$T^{\downarrow^N\uparrow_\pi} =\left( \begin{array}{c}
		1\\
		0.375\\
		0.375\\
		0.375\\
		0.375\\
	\end{array}
	\right)\raisebox{30 pt}{$\uparrow_\pi$}= \left( \begin{array}{c}
		0.5 \\
		0.375 \\
		0.375 \\
		0.375 \\
		0.5 \\
	\end{array}
	\right)=T$$	
	
	The procedures related in~\cite{TFS:2020-acmr, Cornejo2017, ar:ins:2015} can also be applied in order to obtain the reducts of  $(U,V,R)$. In this case, there are two possible reducts: $Y_1=\{u_1, u_2, u_3\}$ and $Y_2=\{u_1, u_3, u_4\}$.
	
	By Corollary~\ref{cor:red-reduct}, two possibilities for reducing FRE~\eqref{exp:eq_max_min} arise. If we fix the reduct $Y_1$, we can solve the FRE
	
	\[R_{Y_1}\odot_\text{G} X =T_{Y_1}\]
	where
	{\fontsize{10}{13}\selectfont $$R_{Y_1}=\left( \begin{array}{ccccc}
		0.5 & 0.25 & 0.75 & 0.675 & 0.25 \\
		0.25 & 0.5 & 0.75 & 0.5 & 0.375 \\
		0.125 & 0.5 & 0.75 & 0.5 & 0.5 \\
	\end{array}\right), \quad T_{Y_1}=
	\left( \begin{array}{c}
		0.5 \\
		0.375 \\
		0.375 \\
	\end{array}
	\right)$$}
\noindent instead of the original one. At this point, the reduced FRE can be solved with any of the procedures that are available in the literature, what will lead to 875 different solutions. They are characterized by a maximum one, 
	\[\overline{X}=\left( \begin{array}{c}
		1\\
		0.375\\
		0.375\\
		0.375\\
		0.375\\
	\end{array}
	\right)\]
	and a set of minimal ones
	\[X_1=\left( \begin{array}{c}
		0.5\\
		0.375\\
		0\\
		0\\
		0\\
	\end{array}
	\right)X_2=\left( \begin{array}{c}
		0.5\\
		0\\
		0.375\\
		0\\
		0\\
	\end{array}
	\right)X_3=\left( \begin{array}{c}
		0.5\\
		0\\
		0\\
		0.375\\
		0\\
	\end{array}
	\right)X_4=\left( \begin{array}{c}
		0.5\\
		0\\
		0\\
		0\\
		0.375\\
	\end{array}
	\right)\]
	Applying Theorem~\ref{th:red_cons_maxmin}, these are the solutions of FRE~\eqref{exp:eq_max_min}.
	
	Notice that, we have easily applied the results introduced in this paper to the particular case of max-min FRE. Similarly, Theorem~\ref{th:approximation_maxmin} can also be applied to the max-min FRE obtaining a new method for approximating this kind of well-known FRE by solvable ones. \hfill\qedsymbol
\end{example}

\section{Conclusions and future work}
Taking into consideration the existing relationship between FRE and concept lattices, this paper has applied for the first time
attribute reduction theory, in the general frameworks of the multi-adjoint property-oriented  concept lattices and object-oriented concept lattices,  in order to simplify FRE, removing redundant equations and preserving the underlying  information in the  original/complete FRE.  
As a consequence, the computation of
the whole solution set of a given FRE is reduced. Indeed, this advantage is even better when a set of FRE with the same coefficient matrix (such as in dynamic systems problems) must be solved. 

Moreover, the introduced FRE reduction method has been used to introduce a procedure for approximating FRE containing some incoherences and usual errors that could be present in real data, in which the existence of uncertainty and imprecision is natural.

In the future, coherence measures of unsolvable FRE will be studied, which can characterize the level of uncertainty and/or imprecision in the considered dataset. Moreover, the introduced approximation mechanism will be combined with the pessimistic and optimistic approaches~\cite{Cornejo2017} to be applied in real cases related to the COST Action DigForASP (digforasp.uca.es).

\end{document}